\documentclass[10pt,twocolumn,letterpaper]{article}

\usepackage{cvpr}
\usepackage{times}
\usepackage{epsfig}
\usepackage{graphicx}
\usepackage{amsmath}
\usepackage{amssymb}
\usepackage{float,wrapfig}
\usepackage{textcomp,booktabs}
\usepackage[usenames,dvipsnames]{color}
\usepackage{colortbl}
\usepackage{color}

\usepackage[pagebackref=true,breaklinks=true,letterpaper=true,colorlinks,bookmarks=false]{hyperref}

\cvprfinalcopy 

\def\cvprPaperID{****} 

\begin{document}

\title{\large \bf Multi-Evidence Filtering and Fusion for Multi-Label Classification, Object Detection and Semantic Segmentation Based on Weakly Supervised Learning}

\author{Weifeng Ge \hspace{0.8in} Sibei Yang \hspace{0.8in} Yizhou Yu\\
Department of Computer Science, The University of Hong Kong
}

\maketitle

\begin{abstract}
Supervised object detection and semantic segmentation require object or even pixel level annotations. When there exist image level labels only, it is challenging for weakly supervised algorithms to achieve accurate predictions. The accuracy achieved by top weakly supervised algorithms is still significantly lower than their fully supervised counterparts. In this paper, we propose a novel weakly supervised curriculum learning pipeline for multi-label object recognition, detection and semantic segmentation. In this pipeline, we first obtain intermediate object localization and pixel labeling results for the training images, and then use such results to train task-specific deep networks in a fully supervised manner. The entire process consists of four stages, including object localization in the training images, filtering and fusing object instances, pixel labeling for the training images, and task-specific network training. To obtain clean object instances in the training images, we propose a novel algorithm for filtering, fusing and classifying object instances collected from multiple solution mechanisms. In this algorithm, we incorporate both metric learning and density-based clustering to filter detected object instances. Experiments show that our weakly supervised pipeline achieves state-of-the-art results in multi-label image classification as well as weakly supervised object detection and very competitive results in weakly supervised semantic segmentation on MS-COCO, PASCAL VOC 2007 and PASCAL VOC 2012.

\end{abstract}

\section{Introduction}
Deep neural networks give rise to many breakthroughs in computer vision by usinging huge amounts of labeled training data. Supervised object detection and semantic segmentation require object or even pixel level annotations, which are much more labor-intensive to obtain than image level labels. On the other hand, when there exist image level labels only, due to incomplete annotations, it is very challenging to predict accurate object locations, pixel-wise labels, or even image level labels in multi-label image classification.

Given image level supervision only, researchers have proposed many weakly supervised algorithms for detecting objects and labeling pixels. These algorithms employ different mechanisms, including bottom-up, top-down~\cite{zhang2016top,lapuschkin2016analyzing} and hybrid approaches~\cite{roy2017combining}, to dig out useful information.
In bottom-up algorithms, pixels are usually grouped into many object proposals,
which are further classified, and the classification results are merged to match groundtruth image labels.
In top-down algorithms, images first go through a forward pass of a deep neural network, and the result is then propagated backward to discover which pixels actually contribute to the final result \cite{zhang2016top,lapuschkin2016analyzing}. There are also hybrid algorithms~\cite{roy2017combining} that consider both bottom-up and top-down cues in their pipeline.

Although there exist many weakly supervised algorithms, the accuracy achieved by top weakly supervised algorithms is still significantly lower than their fully supervised counterparts. This is reflected in both the precision and recall of their results. In terms of precision, results from weakly supervised algorithms contain much more noise and outliers due to indirect and incomplete supervision. Likewise, such algorithms also achieve much lower recall because there is insufficient labeled information for them to learn comprehensive feature representations of target object categories. However, different types of weakly supervised algorithms may return different but complementary subsets of the ground truth.

These observations motivate an approach that first collect as many evidences and results as possible from multiple types of solution mechanisms, put them together, and then remove noise and outliers from the fused results using powerful filtering techniques. This is in contrast to deep neural networks trained from end to end. Although this approach needs to collect results from multiple separately trained networks, the filtered and fused evidences are eventually used for training a single network used for the testing stage. Therefore, the running time of the final network during the testing stage is still comparable to that of state-of-the-art end-to-end networks.


According to the above observations, we propose a weakly supervised curriculum learning pipeline for object recognition, detection and segmentation. At a high level, we obtain object localization and pixelwise semantic labeling results for the training images first using their image level labels, and then use such intermediate results to train object detection, semantic segmentation, and multi-label image classification networks in a fully supervised manner.

Since image level, object level and pixel level analysis has mutual dependencies, they are not performed independently but organized into a single pipeline with four stages.
In the first stage, we collect object localization results in the training images from both bottom-up and top-down weakly supervised object detection algorithms. In the second stage, we incorporate both metric learning and density-based clustering to filter detected object instances. In this way, we obtain a relatively clean and complete set of object instances. Given these object instances, we further train a single-label object classifier, which is applied to all object instances to obtain their final class labels. Third, to obtain a relatively clean pixel-wise probability map for every class and every training image, we fuse the image level attention map, object level attention maps and an object detection heat map. The pixel-wise probability maps are used for training a fully convolutional network, which is applied to all training images to obtain their final pixel-wise label maps. Finally, the obtained object instances and pixel-wise label maps for all the training images are used for training deep networks for object detection and semantic segmentation respectively. To make pixel-wise label maps of the training images help multi-label image classification, we perform multi-task learning by training a single deep network with two branches, one for multi-label image classification and the other for pixel labeling. Experiments show that our weakly supervised curriculum learning system is capable of achieving state-of-the-art results in multi-label image classification as well as weakly supervised object detection and very competitive results in weakly supervised semantic segmentation on MS-COCO~\cite{lin2014microsoft}, PASCAL VOC 2007 and PASCAL VOC 2012~\cite{everingham2010pascal}.


In summary, this paper has the following contributions.

{\flushleft $\bullet$} We propose a novel weakly supervised pipeline for multi-label object recognition, detection and semantic segmentation. In this pipeline, we first obtain intermediate labeling results for the training images, and then use such results to train task-specific networks in a fully supervised manner.
{\flushleft $\bullet$} To obtain clean object instances detected in the training images, we propose a novel algorithm for filtering, fusing and classifying object instances collected from multiple solution mechanisms. In this algorithm, we incorporate both metric learning and density-based clustering to filter detected object instances.
{\flushleft $\bullet$} To obtain a relatively clean pixel-wise probability map for every class and every training image, we propose a novel algorithm for fusing image level and object level attention maps with an object detection heat map. The fused maps are used for training a fully convolutional network for pixel labeling.

\begin{figure*}[ht]
  \centering
  \includegraphics[width=1.0\linewidth]{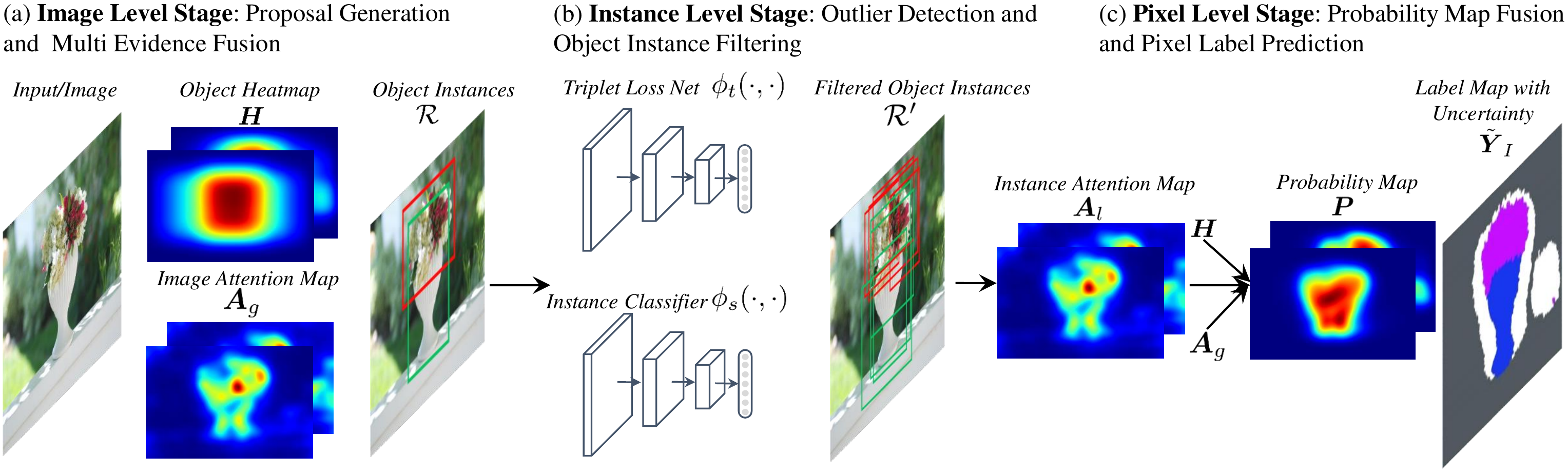}
  \caption{The proposed weakly supervised pipeline. From left to right: (a) Image level stage: fuse the object heatmaps $\boldsymbol{H}$ and the image attention map $\boldsymbol{A}_g$ to generate object instances $\mathcal{R}$ for the instance level stage, and provide these two maps for information fusion at the pixel level stage. (b) Instance level stage: perform triplet loss based metric learning and density based clustering for outlier detection, and train a single label instance classifier $\phi_{s}(\cdot,\cdot)$ for instance filtering. (c) Pixel level stage: integrate the object heatmaps $\boldsymbol{H}$, instance attention map $\boldsymbol{A}_l$, and image attention map $\boldsymbol{A}_g$ for pixel labeling with uncertainty.
  }
  \label{Fig:WSCL}
\end{figure*}

\section{Related Work}
\noindent\textbf{Weakly Supervised Object Detection and Segmentation.} Weakly supervised object detection and segmentation respectively locates and segments objects with image-level labels only~\cite{oquab2015object,diba2016weakly}.
They are important for two reasons: first, learning complex visual concepts from image level labels is one of the key components in image understanding; second, fully supervised deep learning is too data hungry.

Weakly supervised object detection/localization is usually performed in a bottom-up manner. Methods in \cite{oquab2015object,durand2016weldon,durand2017wildcat} treat the weakly supervised localization problem as an image classification problem, and obtain object locations in specific pooling layers of their networks. Methods in \cite{bilen2016weakly,tang2017multiple} extract object instances from images using selective search~\cite{uijlings2013selective} or edge boxes~\cite{zitnick2014edge}, convert the weakly supervised detection problem into a multi-instance learning problem~\cite{dietterich1997solving}. The method in \cite{dietterich1997solving} at first learns object masks as in \cite{durand2016weldon,durand2017wildcat}, and then uses the E-M algorithm to force the network to learn object segmentation masks obtained at previous stages. 
Since it is very hard for a network to directly learn object locations and pixel labels without sufficient supervision, in this paper, we decompose object detection and pixel labeling into multiple easier problems, and solve them progressively in multiple stages.
\vspace{-0mm}

\noindent\textbf{Neural Attention.} Neural attention aims to find out the relationship between the pixels in the input image and the neural responses in every layer of a network. Many efforts~\cite{zhang2016top,bau2017network,lapuschkin2016analyzing} have been made to explain how neural networks work. The method in \cite{lapuschkin2016analyzing} extends layer-wise relevance propagation (LRP)~\cite{bach2015pixel} to comprehend inherent structured reasoning of deep neural networks. To further ignore the cluttered background, a positive neural attention back-propagation scheme, called excitation back-propagation (Excitation BP), is introduced in \cite{zhang2016top}. 
The method in \cite{bau2017network} locates top activations in each convolutional map, and maps these top activation areas into the input image using bilinear interpolation.

\vspace{-0mm}
Neural attention provides a top-down mechanism to obtain pixel-wise class probabilities using image level labels only. In our pipeline, we adopt the excitation BP \cite{zhang2016top} to calculate pixel-wise class probabilities. However for images with multiple category labels, a deep neural network could fuse the activations of different categories in the same neurons. To solve this problem, we train a single-label object instance classification network and perform excitation BP in this network to obtain more accurate pixel level class probabilities.

\noindent\textbf{Curriculum Learning.} Curriculum learning \cite{bengio2009curriculum} is part of the broad family of machine learning methods that starts with easier subtasks and gradually increases the difficulty level of the tasks. In \cite{bengio2009curriculum}, Yoshua {\em et al.} describe the concept of curriculum learning, and use a toy classification problem to show the advantage of decomposing a complex problem into several easier ones. In fact, the idea behind curriculum learning has been widely used before \cite{bengio2009curriculum}. Hinton {\em et al.}~\cite{hinton2006fast} trained a deep neural network layer by layer using a restricted Boltzmann machine~\cite{smolensky1986information} to avoid the local minima in deep neural networks. Many machine learning algorithms~\cite{sun2015robust,graves2017automated} follow a similar divide-and-conquer strategy in curriculum learning.

\vspace{-0mm}
In this paper, we adopt this strategy to decompose the pixel labeling problem into image level learning, object instance level learning and pixel level learning. All the learning tasks in these three stages are relatively simple using the training data in the current stage and the output from the previous stage.

\section{Weakly Supervised Curriculum Learning}

\subsection{Overview}

Given an image $\boldsymbol{I}$ associated with an image level label vector $\boldsymbol{y}_I=[y^1, y^2, ..., y^\mathcal{C}]^T$, our weakly supervised curriculum learning aims to obtain pixel-wise labels $\boldsymbol{Y}_I = [\boldsymbol{y}_1, \boldsymbol{y}_2, ..., \boldsymbol{y}_P]^T$, and then use these labels to assist weakly supervised object detection, semantic segmentation and multi-label image classification.
Here $\mathcal{C}$ is the total number of object classes, $P$ is the total number of pixels in $\boldsymbol{I}$, and $y^l$ is binary. $y^l = 1$ means the $l$-th object class exists in $\boldsymbol{I}$, and $y^l = 0$ otherwise. The label of a pixel $p$ is denoted by a $\mathcal{C}$-dimensional binary vector $\boldsymbol{y}_p$. The number of object classes existing in $\boldsymbol{I}$, which is the same as the number of positive components of $\boldsymbol{y}_I$ is denoted by $K$. Following the divide-and-conquer idea in curriculum learning \cite{bengio2009curriculum}, we decompose the pixel labeling task into three stages: the image level stage, the instance level stage and the pixel level stage.

\subsection{Image Level Stage}

\begin{figure}
  \centering
  \includegraphics[width=1.0\linewidth]{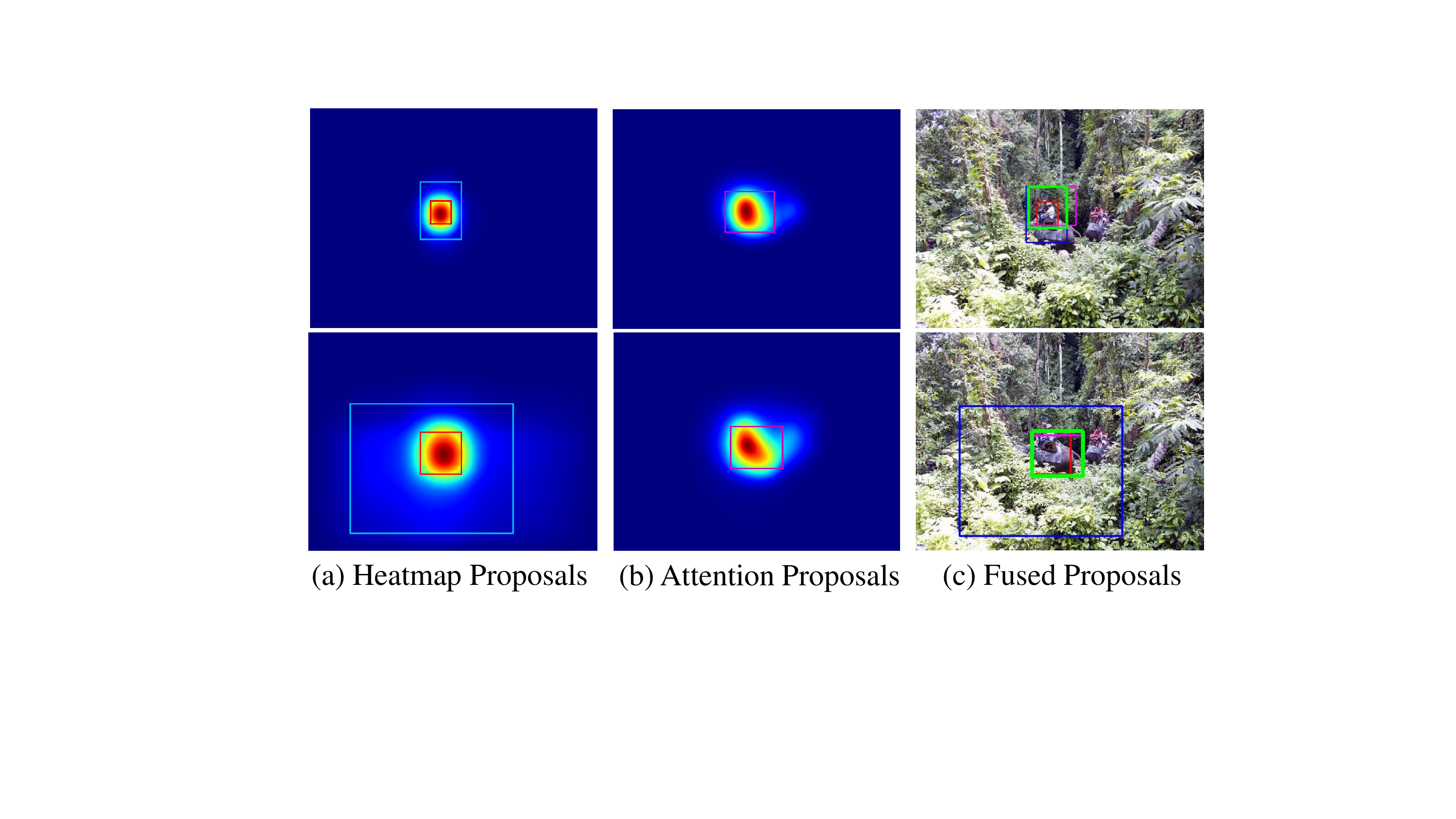}
  \caption{(a) Proposals $\boldsymbol{R}^h$ and $\boldsymbol{R}^l$ generated from an object heatmap, (b) proposals generated from an attention map, (c) filtered proposals (green), heatmap proposals (red and blue), and attention proposals (purple).}
  \label{Fig:Image Level Stage}
\end{figure}

The image level stage not only decomposes multi-label image classification into a set of single-label object instance classifications, but also provides an initial set of pixel-wise probability maps for the pixel level stage.
\vspace{-5mm}
\paragraph{Object Heatmaps.} Unlike the fully supervised case, weakly supervised object detection produces object instances with higher uncertainty and also misses a higher percentage of true objects. To reduce the number of missing detections, we propose to compute an object heatmap $\boldsymbol{H}$ for every object class existing in the image.

For an image $\boldsymbol{I}$ with width $W$ and height $H$, a dense set of object proposals $\boldsymbol{R} = (R_1, R_2, ..., R_n)$ are generated using sliding anchor windows. And the feature stride $\lambda_s$ is set to $8$. The number of locations in the input image where we can place anchor windows is $H/\lambda_s \times W/\lambda_s$. Denote the short side of image $\boldsymbol{I}$ by $L\rho$. Following the setting used for RPN~\cite{ren2015faster}, we let the anchor windows at a single location have four scales $[L\rho/8, L\rho/4, L\rho/2, L\rho]$ and three aspect ratios [0.5, 1, 2]. After proposals out of image borders have been removed, there are usually 12000 remaining proposals per image. Here we define a stack of object heatmaps $\boldsymbol{H} = [\boldsymbol{H}^1, \boldsymbol{H}^2, ..., \boldsymbol{H}^\mathcal{C}]$ as a $\mathcal{C} \times H \times W$ matrix, and all values are set to zero initially. The object detection and classification network $\phi_{d}(\cdot, \cdot)$ used here is the weakly supervised object testing net VGG-16 from \cite{tang2017multiple}. For every proposal $R_i$ in $\boldsymbol{R}$, its object class probability vector $\phi_{d}(\boldsymbol{I}, R_i)$ is added to all the pixels in the corresponding window in the heatmaps. Then every heatmap is normalized to [0, 1] as follows,
    \begin{equation*}
    \begin{aligned}
      \boldsymbol{H}^c = (\boldsymbol{H}^c-min(\boldsymbol{H}^c))/max(\boldsymbol{H}^c),
    \end{aligned}
    \label{eq:objectness heatmap}
    \end{equation*}
where $\boldsymbol{H}^c$ is the heatmap for the $c$-th object class. Note that only the heatmaps for object classes existing in $\boldsymbol{I}$ are normalized. All the other heatmaps are ignored and set to zeros.
\vspace{-5mm}
\paragraph{Multiple Evidence Fusion.} The object heatmaps highlight the regions that may contain objects even when the level of supervision is very weak. However, since they are generated using sliding anchor windows at multiple scales and aspect ratios, they tend to highlight pixels near but outside true objects, as shown in Fig~\ref{Fig:Image Level Stage}. Given an image classification network trained using the image level labels (here we use GoogleNet V1~\cite{zhang2016top}), neural attention calculates the contribution of every pixel to the final classification result. It tends to focus on the most influential regions but not necessarily the entire objects. Note that false positive regions may occur during excitation BP~\cite{zhang2016top}. To obtain more accurate object instances, we integrate the top-down attention maps $\boldsymbol{A}_g = [\boldsymbol{A}^1_g, \boldsymbol{A}^2_g, ..., \boldsymbol{A}^\mathcal{C}_g]$ with the object heatmaps $\boldsymbol{H} = [\boldsymbol{H}^1, \boldsymbol{H}^2, ..., \boldsymbol{H}^\mathcal{C}]$.

For object classes existing in image $\boldsymbol{I}$, their corresponding heatmaps $\boldsymbol{H}$ and attention maps $\boldsymbol{A}_g$ are thresholded by distinct values. The heatmaps $\boldsymbol{H}$ are too smooth to indicate accurate object boundaries, but they provide important spatial priors to constrain object instances obtained from the attention maps. We assume that regions with a sufficiently high value in the object heatmaps should at least include parts of objects, and regions with sufficiently low values everywhere do not contain any objects. Following this assumption, we threshold the heatmaps with two values 0.65 and 0.1 to identify highly confident object proposals $\boldsymbol{R}^h = (R_1^h,R_2^h,...,R_{N_h}^h)$ and relatively low confident object proposals $\boldsymbol{R}^l = (R_1^l,R_2^l,...,R_{N_l}^l)$ after connected component extraction. Then the attention maps are thresholded by 0.5 to attention proposals $\boldsymbol{R}^a = (R_1^a,R_2^a,...,R_{N_a}^a)$ as shown in Fig~\ref{Fig:Image Level Stage}. $N_h$, $N_l$ and $N_a$ are the proposal numbers of $\boldsymbol{R}^h$, $\boldsymbol{R}^l$ and $\boldsymbol{R}^a$. All these object proposals have corresponding class labels. During the fusion, for each object class, the attention proposals $\boldsymbol{R}^a$ which cover more than 0.5 of any proposals in $\boldsymbol{R}^h$ are preserved. We denote these proposals by $\mathcal{R}$, each of which is modified slightly to completely enclose the corresponding proposal in $\boldsymbol{R}^h$ meanwhile be completely contained inside the corresponding proposal in $\boldsymbol{R}^l$ (Fig~\ref{Fig:Image Level Stage}).

\subsection{Instance Level Stage}\label{sec:instance}

\begin{figure}
  \centering
  \includegraphics[width=1.0\linewidth]{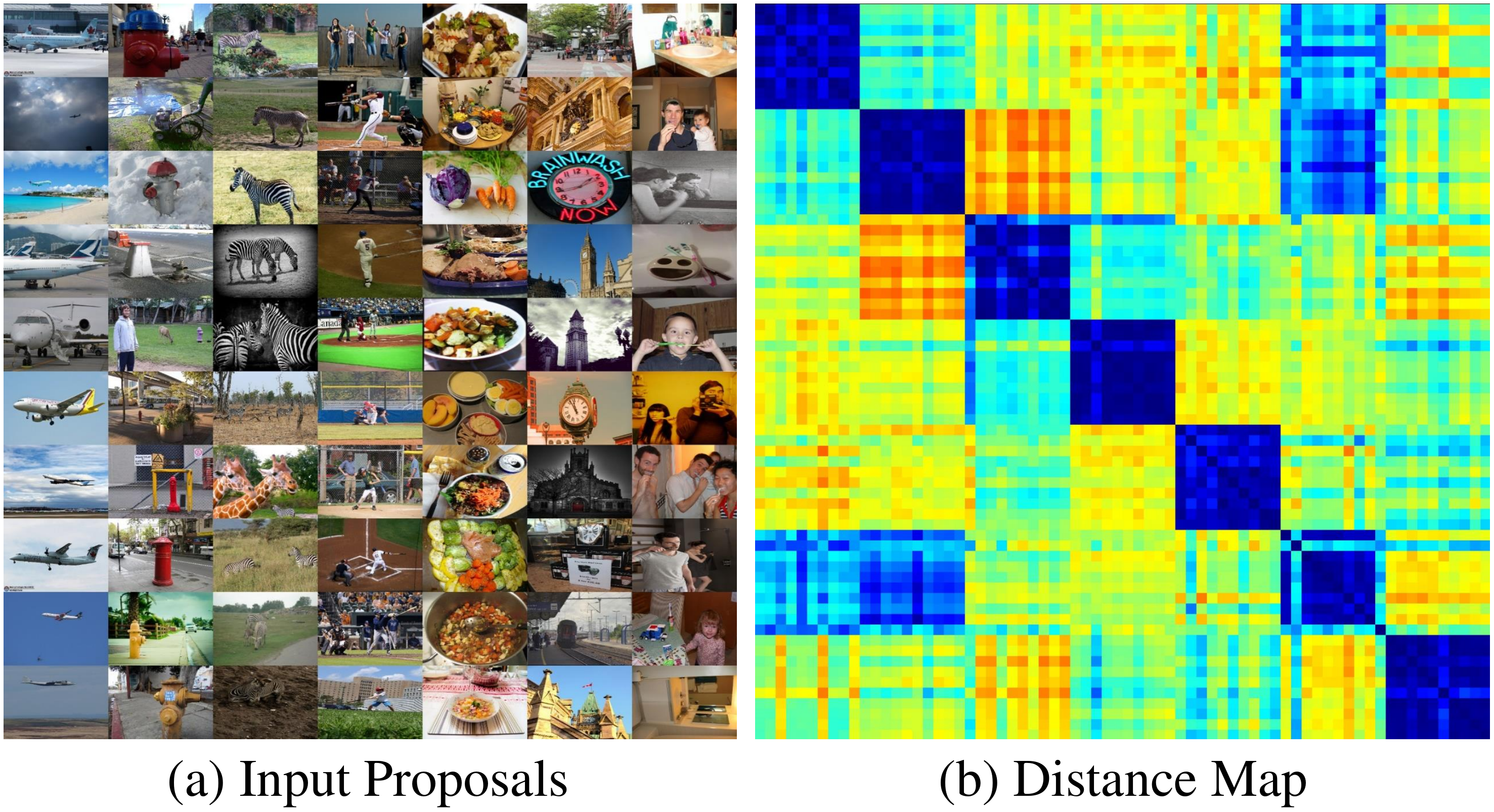}
  \caption{(a) Input proposals of the triplet-loss network, (b) distance map computed using features from the triplet-loss network.}
  \label{Fig:Instance Level Stage}
\end{figure}

Since multiple object categories present in the same image make it hard for neural attention to obtain an accurate pixel-wise attention map for each class, we train a single-label object instance classification network and compute attention maps in this network to obtain more accurate pixel level class probabilities. The fused object instances from the image level stage are further filtered by metric learning and density-based clustering. The remaining labeled object proposals are used for training this object instance classifier, which can also be used to further remove remaining false positive object instances.
\vspace{-4mm}
\paragraph{Metric Learning for Feature Embedding.} Metric learning is popular in face recognition~\cite{schroff2015facenet}, person re-identification and object tracking~\cite{schroff2015facenet,zheng2016person,tsagkatakis2011online}. It embeds an image $\boldsymbol{X}$ into a multi-dimensional feature space by associating this image with a fixed size vector, $\phi_{t}(\boldsymbol{X},\cdot)$, in the feature space. This embedding makes similar images close to each other and dissimilar images apart in the feature space. Thus the similarity between two images can be measured by their distance in this space. The triplet-loss network $\phi_{t}(\cdot, \cdot)$ proposed in \cite{schroff2015facenet} has the additional property that it can well separate classes even when intra-class distances have large variations. When there exist training samples associated with incorrect class labels, the loss stays at a high value and the distances between correctly labeled and mislabeled samples remain very large even after the training process has run for a long time.
Now let $\mathcal{R} = [R_1, R_2, ..., R_O]^T$ denote the fused object instances from all training images in the image level stage, and $\mathcal{Y} = [\boldsymbol{y}_1, \boldsymbol{y}_2, ..., \boldsymbol{y}_O]^T$ are their labels. Here $O$ is the total number of fused instances, and $\boldsymbol{y}_l$ is the label vector of instance $R_l$. We train a triplet-loss network $\phi_{t}(\cdot,\cdot)$ using GoogleNet V2 with BatchNorm as in \cite{schroff2015facenet}. Each mini-batch first chooses $b$ object classes randomly, and then chooses $a$ instances from these classes randomly. These instances are cropped out from the training images and fed into $\phi_{t}(\cdot,\cdot)$. Fig.~\ref{Fig:Instance Level Stage} visualizes a mini-batch composition and the corresponding pairwise distances among instances.
\vspace{-4mm}
\paragraph{Clustering for Outlier Removal.} Clustering aims to remove outliers that are less similar to other object instances in the same class. Specifically, we perform density based clustering~\cite{rodriguez2014clustering} to form a single cluster of normal instances within each object class independently, and instances outside this cluster are considered outliers. This is different from that in \cite{rodriguez2014clustering}. Let $\mathcal{R}^c$ denote instances in $\mathcal{R}$ with class label $c$, and $N_c$ is the number of instances in $\mathcal{R}^c$. Calculate the pairwise distances $d(\cdot, \cdot)$ among these instances, and obtain the $N_c$ by $N_c$ distance matrix $\boldsymbol{D}^c$. For an instance $\mathcal{R}_n^c$, if its distance from another instance is less than $\lambda_d$ (= 0.8), its density $d_n^c$ is increased by 1. Rank these instances by their densities in a descending order, and choose the instance ranked at the top as the seed of the cluster. Then add instances to the cluster following the descending order if their distance to any element in the cluster is less than $\lambda_d$ and their density is higher than $N_c/4$.
\vspace{-4mm}
\paragraph{Instance Classifier for Re-labeling.} Since metric learning and clustering screen object instances in an aggressive way and may heavily decrease their recall, we use the normal instances surviving the previous clustering step to train an instance classifier, which is in turn used to re-label all object proposals generated in the image level stage again.
This is a single-label classification problem as each object instance is allowed a single label. GoogleNet V1 with the SoftMax loss serves as the classifier $\phi_{s}(\cdot,\cdot)$, and it is fine-tuned from the image level classifier. For every object proposal generated in the previous image level stage, if its label predicted by the instance classifier does not match its original label, it is labeled as an outlier and permanently discarded.

\subsection{Pixel Level Stage}

\begin{figure*}[ht]
  \centering
  \includegraphics[width=1.0\linewidth]{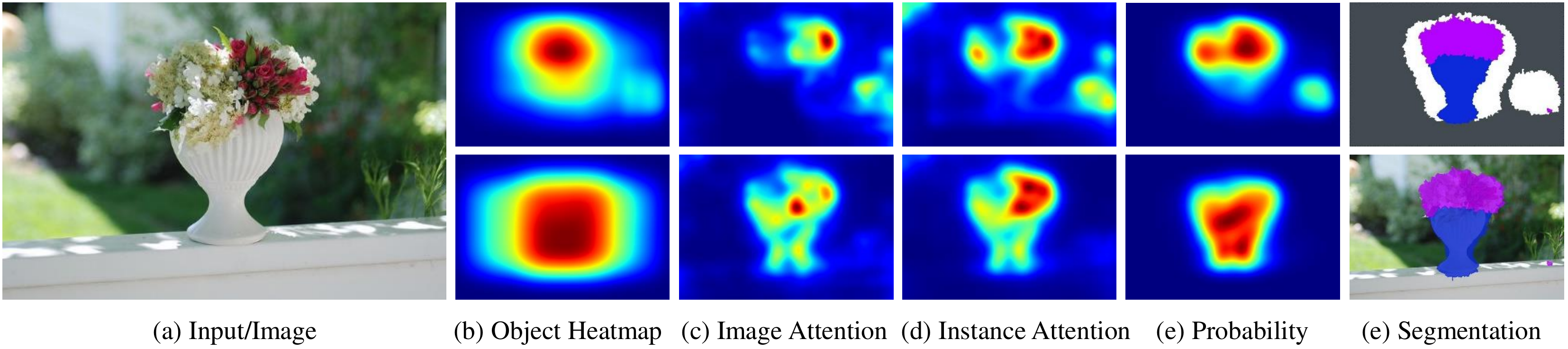}
  \caption{The pixel labeling process in the pixel level stage. White pixels in the last column indicate pixels with uncertain labels.\vspace{-2mm}}
  \label{Fig:Pixel Level Stage}
\end{figure*}

In previous stages, we have already built an image classifier, a weakly supervised object detector, and an object instance classifier. Each of these deep networks produces its own inference result from the input image. For example, the image classifier generates a global attention map, and the object detector generates the object heatmaps. In the pixel level stage, we still perform multi-evidence filtering and fusion to integrate the inference results from all these component networks to obtain the pixelwise probability map indicating potential object categories at every pixel. The global attention map $\boldsymbol{A}_g$ from the image classifier has a full knowledge about the objects in an image but sometimes only focuses on the most important object parts. The object instance classifier has a local view of each individual object. With the help of object-specific local attention maps generated from the instance classifier, we can avoid missing small objects.

\noindent\textbf{Instance Attention Map.} Here we define the instance attention map $\boldsymbol{A}_l$ as a $\mathcal{C} \times H \times W$ matrix, and all values are zero initially. For every surviving object instance from the instance level stage, the object instance classifier $\phi_{s}(\cdot,\cdot)$ is used to extract its local attention map, and add it to the corresponding region in the instance attention map $\boldsymbol{A}_l$. Normalize the range of $\boldsymbol{A}_l$ to [0, 1] as we did for object heatmaps.

\noindent\textbf{Probability Map Integration.} The final attention map $\boldsymbol{A}$ is obtained by calculating the element-wise maximum between the image attention map $\boldsymbol{A}_g$ and the instance attention map $\boldsymbol{A}_l$. That is, $\boldsymbol{A} = max(\boldsymbol{A}_l, \boldsymbol{A}_g)$. For both the heatmap $\boldsymbol{H}$ and the attention map $\boldsymbol{A}$, only the classes existing in the image are considered. The background maps of $\boldsymbol{A}$ and $\boldsymbol{H}$ are defined as follows,
    \begin{equation*}
    \begin{aligned}
      &\boldsymbol{A}_0 = max(0, 1 - \Sigma_{l=1}^\mathcal{C}{{y^l}}\boldsymbol{A}_l),\\
      &\boldsymbol{H}_0 = max(0, 1 - \Sigma_{l=1}^\mathcal{C}{{y^l}}\boldsymbol{H}_l).
    \end{aligned}
    \label{eq:background map}
    \end{equation*}

Now both $\boldsymbol{A}$ and $\boldsymbol{H}$ become $(\mathcal{C}+1) \times H \times W$ matrices. For the $l$-th channel, if $y^l=0$, $\boldsymbol{A}_l = 0$ and $\boldsymbol{H}_l = 0$. Then we perform softmax on both maps along the channel dimension independently. The final probability map $\boldsymbol{P}$ is defined as the result of applying softmax to the element-wise product between $\boldsymbol{A}$ and $\boldsymbol{H}$ by treating $\boldsymbol{H}$ as a filter. That is, $\boldsymbol{P} = \mbox{softmax}(\boldsymbol{H} \odot \boldsymbol{A})$.

\noindent\textbf{Pixel Labeling with Uncertainty.} Pixel labels $\boldsymbol{Y}_I$ are initialized with the probability map $\boldsymbol{P}$. For every pixel $p$, if the maximum element in its label vector $\boldsymbol{y}_p$ is larger than a threshold (=0.6), we simply set the maximum element to 1 and other elements to 0; otherwise, the class label at $p$ is uncertain.

\section{Object Recognition, Detection and Segmentation}

\subsection{Semantic Segmentation}\label{sec:segmentation}
Given pixel-wise labels generated at the end of the pixel level stage for all training images, we train a fully convolutional network (FCN) similar to the network in \cite{long2015fully} to perform semantic segmentation. Note that all pixels with uncertain class labels are excluded during training.
In the prediction part, we adopt atrous spatial pyramid pooling as in \cite{chen2017rethinking}. The resulting trained network can be used for labeling all pixels in any testing image as well as pixels with uncertain labels in all training images.

\subsection{Object Detection}\label{sec:detection}\vspace{-2mm}
Once all pixels with uncertain labels in the training images have been re-labeled using the above network for semantic segmentation, we generate object instances in these images by computing bounding boxes of connected pixels sharing the same semantic label. As in \cite{tang2017multiple} and \cite{li2016weakly}, we train fast RCNN~\cite{girshick2015fast} using these bounding boxes and their associated labels. Since the bounding boxes generated from the semantic label maps may contain noise, we filter them using our object instance classifier as in Section~\ref{sec:instance}. VGG-16 is still the base network of our object detector, which is trained with five scales and flip as in \cite{tang2017multiple}.
\vspace{-2mm}

\subsection{Multi-label Classification}\label{sec:classification}\vspace{-2mm}
The main component in our multi-label classification network is the structure of ResNet-101~\cite{he2016deep}.
There are two branches after layer $res4b22\_relu$ of the main component, one branch for classification and the other for semantic segmentation. Both branches share the same structure after layer $res4b22\_relu$.  Here we adopt multi-task learning to train both branches. The idea is using the training data for the segmentation branch to make the convolutional kernels in the main component more discriminative and powerful.
This network architecture is shown in the supplemental materials. Layer $pool5$ of ResNet-101 in the classification branch is removed, and the output $\boldsymbol{X} (\in \mathbb{R}^{14\times14\times{2048}})$ of layer $res5c$ is a $14\times14\times{2048}$ matrix. $\boldsymbol{X}$ is directly fed into a ${2048\times1\times1\times{C}}$ convolutional layer, and a classification map $\hat{\boldsymbol{Y}}_{cls} (\in \mathbb{R}^{14\times14\times{\mathcal{C}}})$ is obtained.  We let the semantic label map $\hat{\boldsymbol{Y}}_{seg} (\in \mathbb{R}^{14\times14\times{\mathcal{C}}})$ play the role of an attention map $\hat{\boldsymbol{Y}}_{att}$ after the summation over each channel of the semantic label map is normalized to 1. The final image level probability vector $\hat{\boldsymbol{y}}$ is the result of spatial average pooling over the element-wise product between $\hat{\boldsymbol{Y}}_{cls}$ and $\hat{\boldsymbol{Y}}_{att}$. Here $\hat{\boldsymbol{Y}}_{att}$ is used to identify important image regions and assign them larger weights.
At the end, the probability vector $\hat{\boldsymbol{y}}$ is fully connected to an output layer, which performs binary classification for each of the $\mathcal{C}$ classes. The cross-entropy loss is used for training the multi-label classification network.
The segmentation branch uses atrous spatial pyramid pooling to perform semantic segmentation, and softmax is applied to enforce a single label per pixel.

\section{Experimental Results}\vspace{-2mm}
\begin{figure*}[ht]
  \centering
  \includegraphics[width=1.0\linewidth]{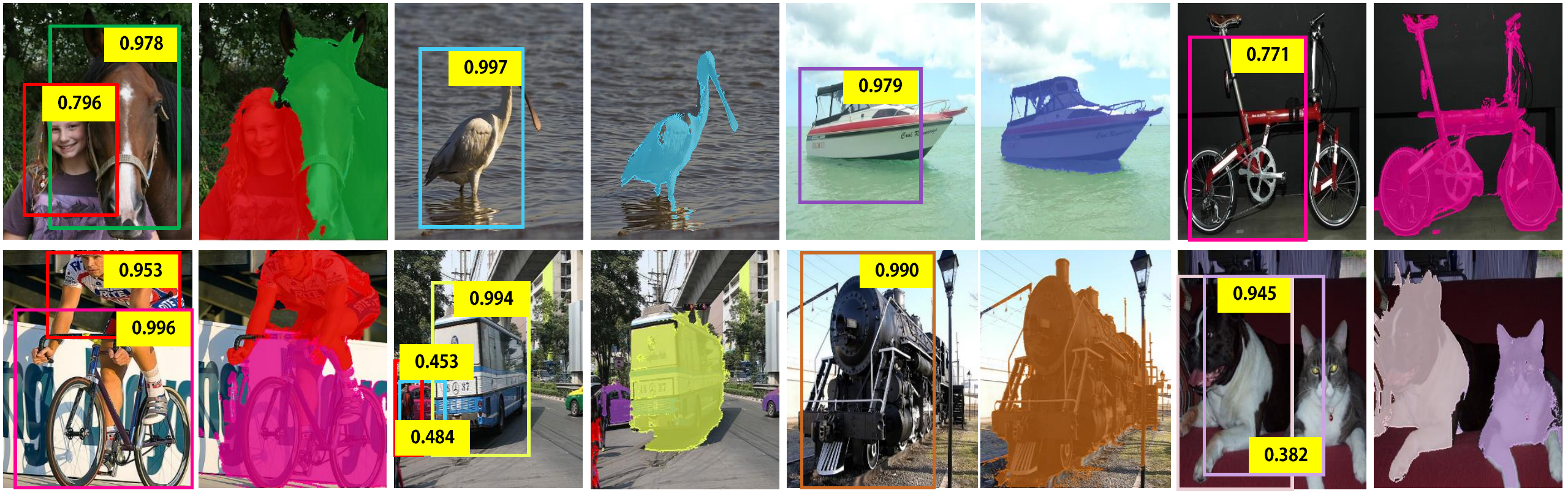}
  \caption{The detection and semantic segmentation results on Pascal VOC 2012 test set (the first row) and Pascal VOC 2007 test set (the second row). The detection results are gotten by select proposals with the highest confidence of every class. The semantic segmentation results are post-processed by CRF \cite{krahenbuhl2011efficient}.}
  \label{Fig:Experimental Results}
\end{figure*}

\begin{table*}[t]\small
\setlength{\abovecaptionskip}{10pt}
\setlength{\belowcaptionskip}{-10pt}
\begin{center}
\resizebox{1\textwidth}{!}
{
\begin{tabular}{@{}lccccccccccccccccccccccc@{}}
\toprule
method                              &bg            &aero          &bike          &bird          &boat          &bottle        &bus           &car           &cat           &chair         &cow           &table         &dog           &horse         &mbike          &person        &plant         &sheep         &sofa          &train         &tv            &mIoU          \\ \midrule
SEC\cite{kolesnikov2016seed}        &83.5          &56.4          &28.5          &64.1          &23.6          &46.5          &70.6          &58.5          &71.3          &\textbf{23.2} &54.0          &28.0          &68.1          &62.1          &70.0           &55.0          &38.4          &58.0          &39.9          &38.4          &48.3          &51.7          \\
FCL\cite{roy2017combining}          &85.7          &58.8          &30.5          &67.6          &24.7          &44.7          &\textbf{74.8} &61.8          &\textbf{73.7} &22.9          &57.4          &27.5          &\textbf{71.3} &64.8          &\textbf{72.4}  &57.3          &37.0          &60.4          &42.8          &42.2          &\textbf{50.6} &53.7          \\
TP-BM\cite{kim2017two}                &83.4          &62.2          &26.4          &\textbf{71.8} &18.2          &\textbf{49.5} &66.5          &\textbf{63.8} &73.4          &19.0          &56.6          &35.7          &69.3          &61.3          &71.7           &\textbf{69.2} &39.1          &66.3          &\textbf{44.8} &35.9          &45.5          &53.8          \\
\cite{Wei_2017_CVPR}                &-          &-          &-          &-  &-          &- &-          &-  &-          &-          &-          &-          &-          &-          &-           &- &-          &-          &-  &-          &-          &55.7           \\ \midrule
Ours+CRF                            &\textbf{86.6} &\textbf{72.0} &\textbf{30.6} &68.0          &\textbf{44.8} &46.2          &73.4          &56.6          &73.0          &18.9          &\textbf{63.3} &32.0          &70.1          &\textbf{72.2} &68.2           &56.1          &34.5          &\textbf{67.5} &29.6          &\textbf{60.2}  &43.6          &\textbf{55.6} \\
\bottomrule
\end{tabular}
}
\end{center}
\caption{Comparison among weakly supervised semantic segmentation methods on PASCAL VOC 2012 $segmentation~test$ set.\vspace{-2mm}}
\label{voc12 test sgementation}
\end{table*}

\begin{table*}[t]\small
\setlength{\abovecaptionskip}{10pt}
\setlength{\belowcaptionskip}{-10pt}
\begin{center}
\resizebox{1\textwidth}{!}
{
\begin{tabular}{@{}lcccccccccccccccccccccc@{}}
\toprule
method                                & aero         &bike  &bird          &boat  &bottle        &bus           &car  &cat           &chair &cow          &table &dog          &horse &mbike         &person &plant         &sheep &sofa         &train &tv                    &mAP   \\ \midrule
OM+MIL+FRCNN\cite{li2016weakly}       & 54.5         &47.4  &41.3          &20.8  &17.7          &51.9          &63.5 &46.1          &21.8  &57.1         &22.1  &34.4         &50.5  &61.8          &16.2   &\textbf{29.9} &40.7  &15.9         &55.3  &40.2                  &39.5  \\
HCP+DSD+OSSH3\cite{jie2017deep}       & 54.2         &52.0  &35.2          &25.9  &15.0          &59.6  &\textbf{67.9}&58.7          &10.1  &\textbf{67.4}&27.3  &37.8         &54.8  &\textbf{67.3} &5.1    &19.7  &\textbf{52.6} &43.5         &56.9  &62.5                  &43.7  \\
OICR-Ens+FRCNN\cite{tang2017multiple} & 65.5         &67.2  &47.2          &21.6  &22.1          &68.0          &68.5 &35.9          &5.7   &63.1         &49.5  &30.3         &64.7  &66.1          &13.0   &25.6          &50.0  &57.1         &60.2  &59.0                  &47.0  \\ \midrule
Ours+FRCNN w/o clustering             & 66.7         &61.8  &55.3          &41.8  &6.7           &61.2          &62.5 &\textbf{72.8} &12.7  &46.2         &40.9  &\textbf{71.0}&67.3  &64.7  &\textbf{30.9}  &16.7          &42.6  &56.0         &65.0  &26.5                  &48.5  \\
Ours+FRCNN w/o uncertainty            & 66.8         &63.4  &54.5          &42.2  &5.8           &60.5          &58.3 &67.8          &7.8   &46.1         &40.3  &71.0         &68.2  &62.6          &30.7   &16.5          &41.1  &55.2         &66.8  &25.2                  &47.5  \\
Ours+FRCNN w/o instances               &\textbf{67.7} &62.9  &53.1  &\textbf{44.4} &11.2          &62.4          &58.5 &71.2          &8.3   &45.7         &41.5  &71.0         &68.0  &59.2          &30.3   &15.0          &42.4  &56.0 &\textbf{67.2} &26.8                  &48.1  \\
Ours+FRCNN                            & 64.3 &\textbf{68.0} &\textbf{56.2} &36.4  &\textbf{23.1} &\textbf{68.5} &67.2 &64.9          &7.1   &54.1 &\textbf{47.0} &57.0 &\textbf{69.3} &65.4          &20.8   &23.2          &50.7  &\textbf{59.6}&65.2  &\textbf{57.0} &\textbf{51.2} \\
\bottomrule
\end{tabular}
}
\end{center}
\caption{Average precision (in \%) of weakly supervised methods on PASCAL VOC 2007 $detection~test$ set.}
\label{voc07 detection}
\end{table*}

\begin{table*}[t]\small
\setlength{\abovecaptionskip}{10pt}
\setlength{\belowcaptionskip}{-10pt}
\begin{center}
\resizebox{0.8\textwidth}{!}
{
\begin{tabular}{@{}lccccccccccccccccccccccc@{}}
\toprule
method                              & F1-C & P-C  & R-C  & F1-O & P-O  & R-O    & F1-C/top3 & P-C/top3 & R-C/top3 & F1-O/top3 & P-O/top3 & R-O/top3 \\ \midrule
CNN-RNN\cite{wang2016cnn}           & -    & -    & -    & -    & -    & -      & 60.4      & 66.0     & 55.6     & 67.8      & 69.2     & 66.4     \\
RLSD\cite{zhang2016multi}           & -    & -    & -    & -    & -    & -      & 62.0      & 67.6     & 57.2     & 66.5      & 70.1     & 63.4     \\
RNN-Attention\cite{wang2017multi}   & -    & -    & -    & -    & -    & -      & 67.4      & 79.1     & 58.7     & 72.0      & 84.0     & 63.0     \\
ResNet101-SRN\cite{zhu2017learning} & 70.0 &\textbf{81.2}& 63.3 & 75.0 & 84.1 & 67.7   & 66.3      &\textbf{85.8}    & 57.5     & 72.1      & 88.1     & 61.1     \\ \midrule
ResNet101($448\times448$)(baseline) & 72.8 & 73.8 &\textbf{72.9}& 76.3 & 77.5 &\textbf{75.1}  & 69.5      & 78.3     &\textbf{63.7}    & 73.1      & 83.8     &\textbf{64.9}    \\
Ours                                &\textbf{74.9}& 80.4 & 70.2 &\textbf{78.4}&\textbf{85.2}& 72.5   &\textbf{70.6}     & 84.5     & 62.2     &\textbf{74.7}     &\textbf{89.1}    & 64.3     \\
\bottomrule
\end{tabular}
}
\end{center}
\caption{Performance comparison among multi-label classification methods on Microsoft COCO 2014 $validation$ set.}
\label{ms coco multilabel}
\end{table*}

All our experiments are implemented using Caffe~\cite{jia2014caffe} and run on an NVIDIA TITAN X(Maxwell) GPU with 12GB memory. The hyper-parameters in Section 3 are set according to common sense and confirmed after we visually verify that the segmentation results on a few training samples are valid. The same parameter setting is used for all datasets and has not been tuned on any validation sets.\vspace{-2mm}
\subsection{Semantic Segmentation}\vspace{-2mm}
\noindent\textbf{Datasets and performance measures.} The Pascal VOC 2012 dataset~\cite{everingham2015pascal} serves as a benchmark in most existing work on weakly-supervised semantic segmentation. It has 21 classes and 10582 training images (the VOC 2012 training set and additional data annotated in \cite{hariharan2011semantic}), 1449 for validation and 1456 for testing. Only image tags are used as training data in our experiments. We report results on both the validation (supplemental materials) and test sets.

\noindent\textbf{Implementation details.} Our network is based on VGG-16. The layers after $relu5\_3$ and layer $pool4$ are removed. Dilations in layers $conv5\_1$, $conv5\_2$, and $conv5\_3$ are set to 2. The feature stride $\lambda_s$ at layer $relu5\_3$ is 8. We add the atrous spatial pyramid pooling as in DeepLab V3 \cite{chen2017rethinking} after layer $relu5\_3$. The dilations in our atrous spatial pyramid pooling layers are $[1, 2, 4, 6]$. This FCN is implemented in py-faster-rcnn \cite{renNIPS15fasterrcnn}. For data augmentation, we use five image scales (480, 576, 688, 864, 1024) (the shorter side is resized to one of these scales) and horizontal flip, and cap the longer side at 1200. During testing, the original size of an input image is preserved. The network is fine-tuned from the pre-trained model for ImageNet in \cite{Simonyan14c}. The learning rate $\gamma$ is set to 0.001 in the first 20k iterations, and 0.0001 in the next 20k iterations. The weight decay is 0.0005, and the mini-batch size is 1. Post-processing using CRF \cite{krahenbuhl2011efficient} is added during testing.

\noindent\textbf{Result comparison.} We compare our method with existing state-of-the-art algorithms.
Table~\ref{voc12 test sgementation} lists the results of weakly supervised semantic segmentation on Pascal VOC 2012. The proposed method achieves 55.6\% mean IoU, comparable to the state of the art (AE-SPL~\cite{Wei_2017_CVPR}). Recent algorithms, including AE-PSL\cite{Wei_2017_CVPR}, F-B~\cite{saleh2016built}, FCL~\cite{roy2017combining}, and SEC~\cite{kolesnikov2016seed}, all conduct end-to-end training to learn object score maps. Our method demonstrates that if we filter and integrate multiple types of intermediate evidences at different granularities during weakly supervised training, the results become equally competitive or even better. 

\subsection{Object Detection}\vspace{-2mm}
\noindent\textbf{Datasets and performance measures.} The performance of our object detector in Section~\ref{sec:detection} is evaluated on the popular Pascal VOC 2007 and Pascal VOC 2012 datasets~\cite{everingham2015pascal}. Each of these two datasets is divided into train, val and test sets. The trainval sets (5011 images for 2007 and 11540 images for 2012) are used for training, and only image tags are used. Two measures are used to test our model: mAP and CorLoc. According to the standard Pascal VOC protocol, the mean average precision (mAP) is used for testing our trained models on the test sets, and the correct localization (CorLoc) is used for measuring the object localization accuracy~\cite{deselaers2012weakly} on the trainval sets whose image tags are already used as training data.

\noindent\textbf{Implementation details.} We use the code for py-faster-rcnn~\cite{renNIPS15fasterrcnn} to implement fast R-CNN~\cite{girshick2015fast}. The network is still VGG-16. The learning rate is set to 0.001 in the first 30k iterations, and 0.0001 in the next 10k iterations. The momentum and weight decay are set to 0.9 and 0.0005 respectively. We follow the same data augmentation setting in \cite{tang2017multiple}, use five image scales (480, 576, 688, 864, 1200) and horizontal flip, and cap the longer image side at 2000.

\noindent\textbf{Result comparison.} Object detection results on Pascal VOC 2007 test set (Table \ref{voc07 detection}) and Pascal VOC 2012 test set (supplemental materials) are reported. Object localization results on Pascal VOC 2007 trainval set and Pascal VOC 2012 trainval set are also reported (supplemental material). On Pascal VOC 2012 test set, our algorithm achieves the highest mAP (47.5\%), at least 5.0\% higher than the latest state-of-the-art algorithms including OICR \cite{tang2017multiple} and HCP+DSD+OSSH3\cite{jie2017deep}. Our trained model also achieves the highest mAP (51.2\%) among all weakly supervised algorithms on Pascal VOC 2007 test set,
4.2\% higher than the latest result from \cite{tang2017multiple}. The object localization accuracy (CorLoc) of our trained model on Pascal VOC 2007 trainval set and Pascal VOC 2012 trainval set are respectively 67\% and 69.4\%, which are 2.7\% and 3.8\% higher than the previous best.

\subsection{Multi-Label Classification}\vspace{-2mm}
\noindent\textbf{Dataset and performance measures.} Microsoft COCO~\cite{lin2014microsoft} is the most popular dataset in multi-label classification. MS-COCO was primarily built for object recognition tasks in the context of scene understanding. The training set is composed of 82081 images in 80 classes, on average 2.9 object labels per image. Since the groundtruth labels of the test set is not available, performance evaluation is conducted on the validation set with 40504 images. We train our models on the training set and test them on the validation set.

Performance measures for multi-label classification is quite different from those for single-label classification. Following \cite{zhu2017learning, wang2017multi}, we employ macro/micro precision, macro/micro recall, and macro/micro F1-measure to evaluate our trained models. For precision, recall and F1-measure, labels with confidence higher than 0.5 are considered positive. ``P-C", ``R-C" and ``F1-C" represent the average per-class precision, recall and F1-measure while ``P-O", ``R-O" and ``F1-O" represent the average overall precision, recall and F1-measure. These measures do not require a fixed number of labels per image. To compare with existing state-of-the-art algorithms, we also report the results of top-3 labels with confidence higher than 0.5 as in \cite{wang2017multi}.

\noindent\textbf{Implementation details.} Our main network for multi-label classification is ResNet-101 as described earlier. The resolution of the input images is at $448\times448$. We first train a network with the classification branch only. As a common practice, a pre-trained model for ImageNet is fine-tuned with the learning rate $\gamma$ set to 0.001 in the first 20k iterations, and 0.0001 in the next 20k iterations. The weight decay is 0.0005. Then we add the segmentation branch and train this new branch only by fixing all the layers before layer $res4b22\_relu$ and the classification branch. The learning rate is set to 0.001 in the frist 20k iterations, and 0.0001 in the next 20k iterations. At last, we train the entire network with both branches using the cross-entropy loss for multi-label classification for 30k iterations with a learning rate 0.0001 while still fixing the layers before layer $res4b22\_relu$.

\noindent\textbf{Result comparison.}  In addition to our two-branch network, we also train a ResNet-101 classification network as our baseline. The multi-label classification performance of both networks on MS-COCO is reported in Table~\ref{ms coco multilabel}. Since the input resolution of our baseline is $448\times448$, in comparison to the latest work (ResNet101-SRN)~\cite{zhu2017learning}, the performance of our baseline is slightly better. Specifically, the F1-C of our baseline is 72.8\%, which is 2.8\% higher than the F1-C of ResNet101-SRN.
In comparison to the baseline, our two-branch network further achieves overall better performance. Specifically, the P-C of our two-branch network is 6.6\% higher than the baseline, the R-C is 2.7\% lower, and the F1-C is 2.1\% higher. All F1-measures (F1-C, F1-O, F1-C/top3 and F1-O/top3) of our two-branch network are the highest among all state-of-the-art algorithms.

\subsection{Ablation Study}
We perform an ablation study on Pascal VOC 2007 detection test set by replacing or removing a single component in our pipeline every time. First, to verify the importance of object instances, we remove all steps related to object instances, including the entire instance level stage and the operations related to the instance attention map in the pixel level stage. The mAP is decreased by 3.1\% as shown in Table~\ref{voc07 detection}. Second, the clustering and outlier detection step in the instance level stage is removed. We directly train an instance classifier using the object proposals from the image level stage. The mAP is decreased by 2.7\%. Third, instead of labeling a subset of pixels only in the pixel level stage, we assign a unique label to every pixel even in the case of low confidence. The mAP drops to 47.5\%, 3.7\% lower than the performance of the original pipeline.
\section{Conclusions}\vspace{-2mm}
In this paper, we have presented a new pipeline for weakly supervised object recognition, detection and segmentation. Different from previous algorithms, we fuse and filter object instances from different techniques and perform pixel labeling with uncertainty. We use the resulting pixel-wise labels to generate groundtruth bounding boxes for object detection and attention maps for multi-label classification. Our pipeline has achieved clearly better performance in all of these tasks. Nevertheless, how to simplify the steps in our pipeline deserves further investigation.

{\small
\bibliographystyle{ieee}
\bibliography{egbib}
}

\end{document}